\documentclass[sigconf]{acmart}

\usepackage{balance}

\def\BibTeX{{\rm B\kern-.05em{\sc i\kern-.025em b}\kern-.08emT\kern-.1667em\lower.7ex\hbox{E}\kern-.125emX}}
    
\copyrightyear{2019} 
\acmYear{2019} 
\setcopyright{acmlicensed}
\acmConference[UMAP '19]{27th Conference on User Modeling, Adaptation and Personalization}{June 9--12, 2019}{Larnaca, Cyprus}
\acmBooktitle{27th Conference on User Modeling, Adaptation and Personalization (UMAP '19), June 9--12, 2019, Larnaca, Cyprus}
\acmPrice{15.00}
\acmDOI{10.1145/3320435.3320471}
\acmISBN{978-1-4503-6021-0/19/06}

\acmSubmissionID{umap085s}

\usepackage{algorithm}
\usepackage{algorithmic}
\usepackage{subcaption}

\usepackage{tikz}

\usetikzlibrary{arrows}
\tikzset{
  treenode/.style = {align=center, inner sep=0pt, text centered,
    font=\sffamily},
  arn_n/.style = {treenode, circle, white, font=\sffamily\bfseries, draw=black,
    fill=black, text width=1.5em},
  arn_r/.style = {treenode, circle, red, draw=red, 
    text width=1.5em, very thick},
  arn_x/.style = {treenode, rectangle, draw=black,
    minimum width=0.5em, minimum height=0.5em, fill=black}
}

\begin{document}

%
\title{Value Driven Representation for Human-in-the-Loop Reinforcement Learning}

%
\author{Ramtin Keramati}
\affiliation{
  \institution{Institute of Computational and Mathematical Engineering}
  \streetaddress{353 Serra Mall}
  \city{Stanford}
  \state{California}
  \postcode{94305}
}
\email{keramati@cs.stanford.edu}

\author{Emma Brunskill}
\affiliation{%
 \institution{Department of Computer Science}
  \streetaddress{353 Serra Mall}
  \city{Stanford}
  \state{California}
  \postcode{94305}
}
\email{ebrun@cs.stanford.edu}
%
\renewcommand{\shortauthors}{}

%
\begin{abstract}
Interactive adaptive systems powered by Reinforcement Learning (RL) have many potential applications, such as intelligent tutoring systems. In such systems there is typically an external human system designer that is creating, monitoring and modifying the interactive adaptive system, trying to improve its performance on the target outcomes. In this paper we focus on algorithmic foundation of how to help the system designer choose the set of sensors or features to define the observation space used by reinforcement learning agent. We present an algorithm, value driven representation (VDR), that can iteratively and adaptively augment the observation space of a reinforcement learning agent so that is sufficient  to  capture a  (near)  optimal  policy. To  do  so  we  introduce  a  new method to optimistically estimate the value of a policy using
offline  simulated  Monte  Carlo  rollouts. We evaluate the performance of our approach on standard RL benchmarks with simulated humans and demonstrate significant improvement over prior baselines.
\end{abstract}

%
%

\ccsdesc[300]{Human-centered computing~Human computer interaction (HCI)}

\copyrightyear{2019} 
\acmYear{2019} 
\setcopyright{acmlicensed}
\acmConference[UMAP '19]{27th Conference on User Modeling, Adaptation and Personalization}{June 9--12, 2019}{Larnaca, Cyprus}
\acmBooktitle{27th Conference on User Modeling, Adaptation and Personalization (UMAP '19), June 9--12, 2019, Larnaca, Cyprus}
\acmPrice{15.00}
\acmDOI{10.1145/3320435.3320471}
\acmISBN{978-1-4503-6021-0/19/06}

%
\keywords{Reinforcement Learning, Human-in-the-Loop}

%

%
\maketitle

\section{INTRODUCTION}

\begin{figure}[tb!]
\centering
  \includegraphics[width=0.9\linewidth]{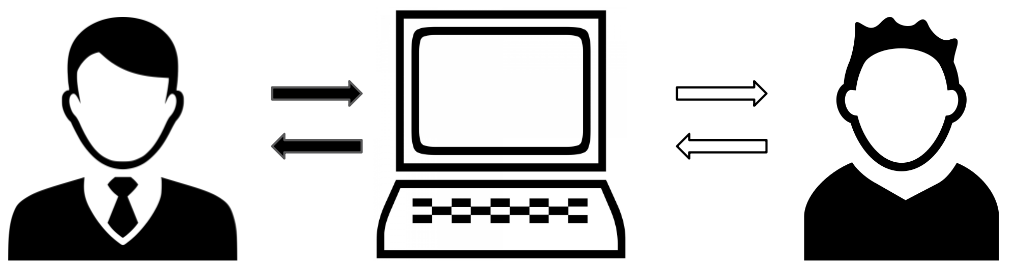}
\caption{We consider interactive adaptive systems which may interact with people (person on the right) and which have a system designer (person on the left) that is trying to optimize the adaptive system. We look at a case where the adaptive system proactively provides input to the designer (outgoing arrow) to propose potential modifications to its sensors that may help improve performance.}
\label{fig:humanloop}

\end{figure}

Interactive adaptive systems powered by reinforcement learning can improve over time and have 
many potential applications. These include intelligent tutoring systems that improve their 
teaching as they instruct more students, smart home devices that adjust temperature settings 
in response to weather and human preferences, and mobile wellness applications that support 
chronic care management. In such cases the system is learning a good decision policy-- 
a mapping from a current observation (of a student, of a home, of a person and their context) to 
action (what pedagogical activity to propose, how to set the temperature, what health nudge to suggest) 
in order to maximize overall outcomes over time (how much student has learned, minimize energy, 
maximize total activity).

In such systems there is typically an external system designer that is creating, monitoring 
and modifying the interactive adaptive system, trying to improve its performance on the 
target outcomes (Figure~\ref{fig:humanloop}).
A key question is how interactive adaptive systems can better provide input back to 
their system designers to help support the designer in improving the performance. 
In this paper we focus on one particular direction for this, how to help the 
system designer choose the set of sensors or features to 
define the observation space used by the interactive system to make decisions. 
For example, does the tutoring system only log student responses to problems, 
or is a webcam also used to detect frustration levels? 

The observation space specification 
implicitly constrains the decision policy class the interactive 
system can optimize over. 
Here we consider the case when the interactive system itself can monitor 
its own performance, and propose to the system designer some potential 
modifications to the observation space that it expects might yield improved 
performance. The human designer can then choose to augment the system 
with additional features and/or sensors that the adaptive system can 
then use going forward. For example, the system may recognize that 
there are number of observations which are currently identical (for example, same 
performance on a set of problems) where selecting the 
action (next problem) does not seem to yield the best outcomes, 
and ask the human if there might be an additional feature/ sensor 
 that 
could be used to distinguish such currently aliased observations. This set of observations 
can provide information to the human designer about what types of 
features/ sensors might be useful to add to the system, like a webcam and emotion classifier to detect frustration. 

In this short paper we focus on the algorithmic foundations of 
this idea, providing a proof of concept in simulated domains. 
Precisely, we present an algorithm, value driven representation (VDR), that can iteratively augment 
the observation space of a reinforcement learning 
agent (such as an interactive adaptive system), if the algorithm 
estimates that the resulting augmented observation space could yield an improved policy performance in the real environment. VDR can be applied both to situations where additional features could be added later in a demand driven way (from a human system designer), or when the full set of features is known in advance but there are computational, performance and interpretability benefits to starting with a more compact representation. 

Our approach starts with a coarse observation space (a small minimal set of features). Note that
this small subset may only be known because these are the initial features seem to be relevant 
by a human system designer or due to cost or other constraints. We 
assume the algorithm is part of an adaptive interactive system that is acting in a 
Markov decision process (MDP), but the set of initial features provided may be a small subset of 
the features needed to satisfy the Markov assumption. We can view the coarse observation space as a state abstraction. A goal of our algorithm is 
to be able to augment the observation space in 
order to reach the minimal set of features sufficient to make the same optimal 
decisions as would be possible with the full (unknown) set of features (formally 
known as a optimal $\pi^*$-irrelevance abstraction~\citep{li2006towards}). 

Our algorithm proceeds by proposing splits of existing observations that look identical under 
the current set of features. The key contribution of our work is to estimate the 
potential value of policies with new augmented observation space using old data without making 
the Markov assumption. Our algorithm for doing so is inspired by Upper Confidence Trees 
\citep{kocsis2006bandit}, a popular Monte Carlo Tree Search method, but adapted to 
focus on decision policy evaluation and, more importantly, does not require an MPD dynamics 
and reward so that it can be run using old data.

We evaluate the performance of our approach on several simulation domains where 
the true Markovian state space is known and demonstrate significant improvement over prior baselines.
While a key next step is to try this out with a human designer in the loop, this is 
an encouraging step of the potential benefit of adaptively adjusting the feature representation.

\section{Related Works}
Reinforcement Learning in non-Markovian observation space has been long studied. UTree~\citep{mccallum1996learning} is a history-based method that uses tree-based representations of the value function and splits observation based on local gain and predictive power. Predictive state representation (PSR)~\citep{littman2002predictive, james2004planning, aberdeen2007policy} is another history-based method that tries to find the sufficient statistics from history to represent a notion of state. In contrast, our algorithm focuses on utility gain (gain in the value of a policy) rather than the predictive power of the state representation. 

Feature RL~\citep{hutter2009feature} is a framework that defines a mapping between history to states such that state representation becomes Markovian, and then uses general RL algorithms to solve the proposed MDP. A brief summary of FRL can be found in~\cite{daswani2014feature}. The main difference of our work with this line of research is that our agent does not seek a Markovian representation and finds the policy in a possibly non-Markovian observation space. Many other related works are based on the AIXI agent~\citep{hutter2004universal} a formal mathematical solution to the general RL agent, e.g. MC-AIXI-CTW~\citep{nguyen2011feature, veness2011monte}
; however, in these methods, the policy representation is not explicit and the agent needs to run UCT at every step. 

Our work strongly relates to the state aggregation/abstraction literature ~\citep{singh1995reinforcement,timmer2006abstract, ravindran2003smdp, anand2015asap}.
However, our work differentiates itself with those in the way that our algorithm starts learning in a small observation space that is often non-Markovian and then trying to augment the observation space to learn the optimal policy, similar to some Bayesian methods like iPOMDP~\citep{doshi2009infinite} which learns a POMDP while growing the state space.

\section{Problem Setup}\label{sec:prob}
We consider human-in-the-loop reinforcement learning, where a human system designer can
modify the observation space definition used by a reinforcement learning agent, such as when 
a designer can modify the observation space of a RL intelligent tutoring system interacting students. 
More precisely, we assume the RL agent is acting in an episodic Markov decision process 
$M = \langle \mathcal{S}, \mathcal{A}, T, R, \gamma \rangle$, where $\mathcal{S}$ is a 
finite set of states (such as a student's current state of learning), $\mathcal{A}$ is a 
finite set of actions, and $T$ is a dynamics model that specifies $p(s'|s,a)$ -- the probability 
of transitioning to state $s'$ after taking action $a$ in state $s$ (for example, the 
probability the student will not understand 1 digit addition, do a problem on addition, 
and transition to a new state $s'$ in which the student understands addition.)
 $R$ specifies the reward $r(s,a)$ received by taking action $a$ in 
state $s$: e.g., high reward when a student takes a test and passes it. $\gamma$ 
is a discount factor that weighs immediate vs future reward.


We assume the state space $\mathcal{S}$ (such as the true internal state of the student) 
is only indirectly observable by the agent through sensors that provide 
the observation space $\mathcal{O}$ (e.g.,the agent can  
observe if the student got a problem correct). 
There is a many-to-one deterministic mapping from states to observations, and 
therefore the observation space can be viewed as an aggregated state space. 
We denote the aggregated states $s$ under observation $o$ as $\mathcal{S}_o$. 

A decision policy $\pi$ for the RL agent is a stochastic mapping from states to 
actions. The state-action value of a policy $Q^{\pi}(s,a)$ is the expected discounted sum of rewards the RL agent would obtain by taking action $a$ from state $s$ and then 
following the policy. In RL the dynamics and reward model are unknown. 


The agent is provided with an initial observation space $\mathcal{O}_0$ that 
can be modified by the human system designer. The goal is for the interactive 
reinforcement learning agent to, together with the designer, find the smallest
observation space that yields the maximal expected reward policy such 
that the resulting policy matches the performance of the best policy under 
the (unknown) MDP. 


\section{Algorithm}

We present a novel human-in-the-loop RL algorithm, Value Driven Representation (VDR). 
VDR involves two key components. First, VDR performs optimistic reinforcement 
learning given the current observation space specification (e.g. it tries to 
optimize a decision policy for teaching a student given the current available features that distinguish 
student learning states) (Section \ref{sec:opto}). Second, VDR evaluates potential augmentations of the 
existing observation space that might enable a better decision policy and 
proposes to a human designer to split the observation that is evaluated to be most beneficial 
(e.g. for the human to provide another feature, like frustration, that can 
be used to refine an observation from "solved problem 1" to "solved problem 1 with 
frustration" and "solved problem 1 without frustration") (Section \ref{sec:aug}). 
Note in this initial work we 
only simulate input human experts and leave a human user study to later work. 

\begin{algorithm}
\caption{VDR}\label{alg:all}
\begin{algorithmic}[1]
{\small
\STATE $O \gets O_{0}$, $\mathcal{D} \gets []$ // data
\FOR{each episode}
\STATE $(\pi,V^{\pi}) \gets$ optimistic Off-Policy Policy Optimization ($\mathcal{D}$)
\STATE Execute $\pi$ for 1 episode and collect trajectory $\tau$, $\mathcal{D} \gets$ $\mathcal{D}\ \cup$  $\tau$ 
\FOR{all observations $o_i \in \mathcal{O}$}
    \STATE Propose augmentation of $o_i \to (o_{i}^1,o_{i}^2)$ // section 3.2
    \STATE $\mathcal{D}' \gets$ relabel $o_i$ in the data $\mathcal{D}$ with $o_{i}^1,o_{i}^2$ 
    \STATE $(\pi_{o_i}^*, V^{\pi_{o_i}^*}) \gets$ Off-Policy Policy Optimization ($\mathcal{D}'$) //section 3.1
\ENDFOR
\STATE ($score_{best}$, $o_{best}$) $\leftarrow \underset{o_i}{\max}\ D_{KL}(\pi^*_{o_i}||\pi) (V^{\pi^*_{o_i}} - V^{\pi})$
\IF{ $score_{best} \geq c$}
    \STATE $O \leftarrow O \cup \{o_{best}^1, o_{best}^2\} \setminus o_{best}$  
\ENDIF
\ENDFOR
}
\end{algorithmic}
\end{algorithm}

\subsection{Off-Policy Policy Optimization}\label{sec:opto}

We first consider how an RL agent should act given the current experience 
in order to quickly learn a good decision policy for the current observation 
space. A key consideration here is that the current observation space 
$\mathcal{O}$ is generally not Markovian: therefore standard techniques 
that rely on the system description being a Markov process (like 
estimating a Markov dynamics and reward model and planning) can (and often will) 
fail~\cite{mccallumThesis,mandel2014offline}.

Instead, we propose an Off-Policy Tree Evaluation (OPTE) approach, 
similar to Monte Carlo Tree Search (MCTS) planning~\cite{kocsis2006bandit} and evaluation methods. 
The key distinction
is that \textit{we do not assume access to a domain model of the decision process} which standard MCTS methods rely on to perform simulations. 
OPTE first uses prior
data to construct a tree representing observed
trajectory sequences, storing counts of the number of times each node has been previously visited at each node. We call this a trajectory tree $\mathcal{T}$.

Similar to MCTS, OPTE uses the tree structure to simulate potential 
sequences of observations, actions and rewards (called rollouts). 
A single rollout involves starting at the
root node ($o_{t_0})$ of $\mathcal{T}$ and sampling transitions using 
a maximum likelihood model of transition probabilities and rewards
given the data associated with this node. It is possible the observed outcomes
have not included all feasible next observations. To
handle this case, our algorithm maintains a pseudo-count
of $C = 1$ over an additional next unseen observation transition, for
each action in the tree. If this outcome is sampled,
or if an action has not been tried from a particular node 
previously (e.g. at step $d$), we terminate the simulated roll-out and use a model free Monte Carlo estimate of the current observation-action
pair Q-value $Q(o_d,a_d)$ at the leaf node, computed by averaging over all returns obtained after 
observing $(o,a)$ tuple. Our complete Off Policy Tree Evaluation (OPTE)
is shown in algorithm~\ref{alg:opte} and \ref{alg:next}, which takes the average of $N$ simulations. $V^{\pi}_{\mathcal{O}} = \frac{1}{N}\sum_{i=1}^{N}{(\sum_{t=1}^{d_i}\gamma^{t-1}r_t + \gamma^{d_i}Q(o_{d_i},a_{d_i}))}$

\begin{algorithm}[tb]
\caption{OPTE($\pi, graph,N, opt$)}\label{alg:opte}
\begin{algorithmic}[1]
{\small
\REQUIRE $\pi$: evaluation policy, $graph$: trajectory tree, $N$: number of simulations, $opt$: optimism flag
\FOR{$i=1:N$}
\STATE $node \gets graph.initialNode()$
\STATE $o \gets node.observation()$; $R(i) \gets []$; $done \gets False$
\WHILE{not $done$}
    \STATE $node,r, done \gets next (node, \pi(o), opt)$
    \STATE $R(i).append(r)$; $o \gets node.observation()$
\ENDWHILE 
\ENDFOR
\RETURN $\frac{1}{N}\sum_i \sum_t \gamma^t R_t(i)$ 
}
\end{algorithmic}
\end{algorithm}

\begin{algorithm}[h!]
\caption{next($node,a,opt$)}\label{alg:next}
\begin{algorithmic}[1]
{\small
\REQUIRE $node$: node in trajectory graph, $a$: action, $opt$: optimism falg
\STATE $C \gets 1$
\STATE $counts \gets node.getTransitionCounts(a)$
\STATE $counts.append(C)$ // Unobserved outcome 
\STATE $model \sim MLE(counts)$ 
\STATE $nextNode \sim model(a)$
\IF{$nextNode$ is -1}
    \STATE // Sample a new potential outcome
	\STATE $o \gets node.OBSERVATION()$; $done \gets True$
	\IF{$opt==True$}
	    \STATE $r \gets graph.Q[o, a] +  \frac{r_{max}}{1-\gamma}     \sqrt{\frac{log(n[o])}{n[o,a]}}$
	\ELSE
	    \STATE $r \gets graph.Q[o, a] $
	\ENDIF
\ELSE
    \STATE $r,done \gets nextNode.Info()$
\ENDIF
\RETURN $(nextNode, r, done)$
}
\end{algorithmic}
\end{algorithm}

Additionally, in order to encourage strategic exploration, we propose a method that is inspired by optimism under uncertainty
approaches.
Precisely, we take the OPTE
algorithm (Algorithm~\ref{alg:opte})  described above and add in a reward
bonus to the $Q(o,a)$ used at the tree leaves.
Similar to upper confidence bound
RL algorithms~\citep{osband2014near} we use 
$Q(o,a) + \frac{r_{max}}{1-\gamma} \sqrt{log(n(o))/n(o,a)}$, 
where $r_{max}$ is the maximum reward and $n(o), n(o,a)$ are visitation counts of an observation and an observation-action pair. 
This approach is a minor modification of OPTE,
which can be computed by setting the input \textit{opt} to \textit{True} in algorithm \ref{alg:next}.
Off-Policy Tree Optimization (OPTO) can be done using any policy optimization method combined with using simulated roll outs on 
the trajectory tree $\mathcal{T}$. 

\subsection{Observation Augmentation}\label{sec:aug}

The initial input observation space may be insufficient to achieve a 
high performance policy: for example, it may be crucial to 
change the policy depending on whether a student is frustrated 
after completing problem 1 correctly, yet initially this distinction 
may be lacking in the observation space. We now propose how 
the RL system can itself try to identify which observation 
refinements might yield an improved performance (policy value) 
if a human designer could provide a feature that distinguished 
between observations that are currently aliased.

To do this, at every episode the algorithm creates and scores  $\{\mathcal{O}_i\}$ new potential observation spaces. 
Each observation space $\mathcal{O}_{i}$ is derived from taking observation $o_i$ in the current $\mathcal{O}_{current}$ 
observation space, splitting it into two new observations $o_{i}^{1}$ and $o_{i}^{2}$, and adding these two new observations to all of the other non-split observations $o_{k \neq i} \in \mathcal{O}_{current}$ ($\{o_{i}^{1}, o_{i}^{2}, \forall_{k \neq i}: o_k \in \mathcal{O}_{current}$\}).

Splitting a particular observation $o_i$ into two is  performed by executing Expectation Maximization (EM) \citep{moon1996expectation}
on the existing collected trajectories to hypothesize 2 potential latent observations with different 
dynamics and/or rewards models. Given the EM learned parameters for a observation split ($o_i$ into observations $o_i^1$ and $o_i^2$), 
the Viterbi algorithm \citep{forney1973viterbi} can be used to relabel prior trajectories,
turning all instances of observation $o_i$ into $o_i^1$ or $o_i^2$. Using the relabeled trajectories $\mathcal{D}'$, we can build a new trajectory tree $\mathcal{T}'$ and perform off-policy policy optimization using OPTO to estimate the value of the best policy $V^{\pi^*_{o_i}}$ for the modified space. 

The objective is to present an augmentation to the system designer if 1) splitting an observation 
yields an optimal policy with a higher value than the existing best policy for the observation space, and 2) if the best policy for the augmented observation (evaluated using OPTO) is sufficiently different than the previous best policy, measured by the KL-divergence of two policies. Precisely, define $score(i) = D_{KL}(\pi^*_{old}||\pi_i)(V^{\pi_i} - V^{\pi_{old}})$
where $D_{KL}$ is the KL-divergence, $\pi_i$ is the optimal policy after splitting observation $o_i$. 
Our agent proposes an observation augmentation $i = argmax_i\ score(i)$ to the system designer if $\max_i score(i) \geq c$ exceeds a threshold $c$.
\begin{figure}[tb]
\centering
\begin{subfigure}{.49\linewidth}
  \centering
  \includegraphics[width=.75\linewidth]{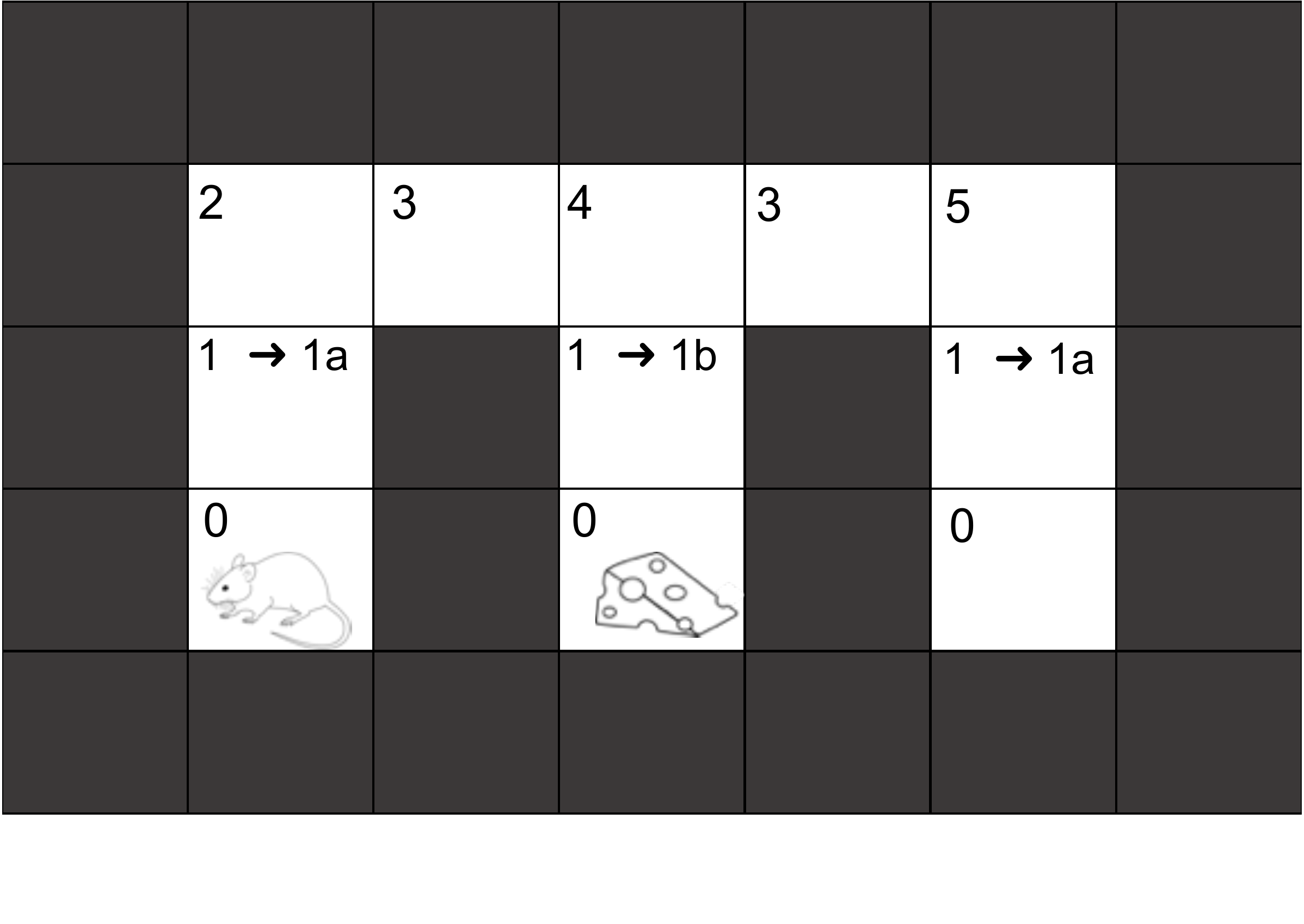}
  \caption{}
  \label{fig:CMlayout}
\end{subfigure}%
\begin{subfigure}{.49\linewidth}
  \centering
  \includegraphics[width=\linewidth]{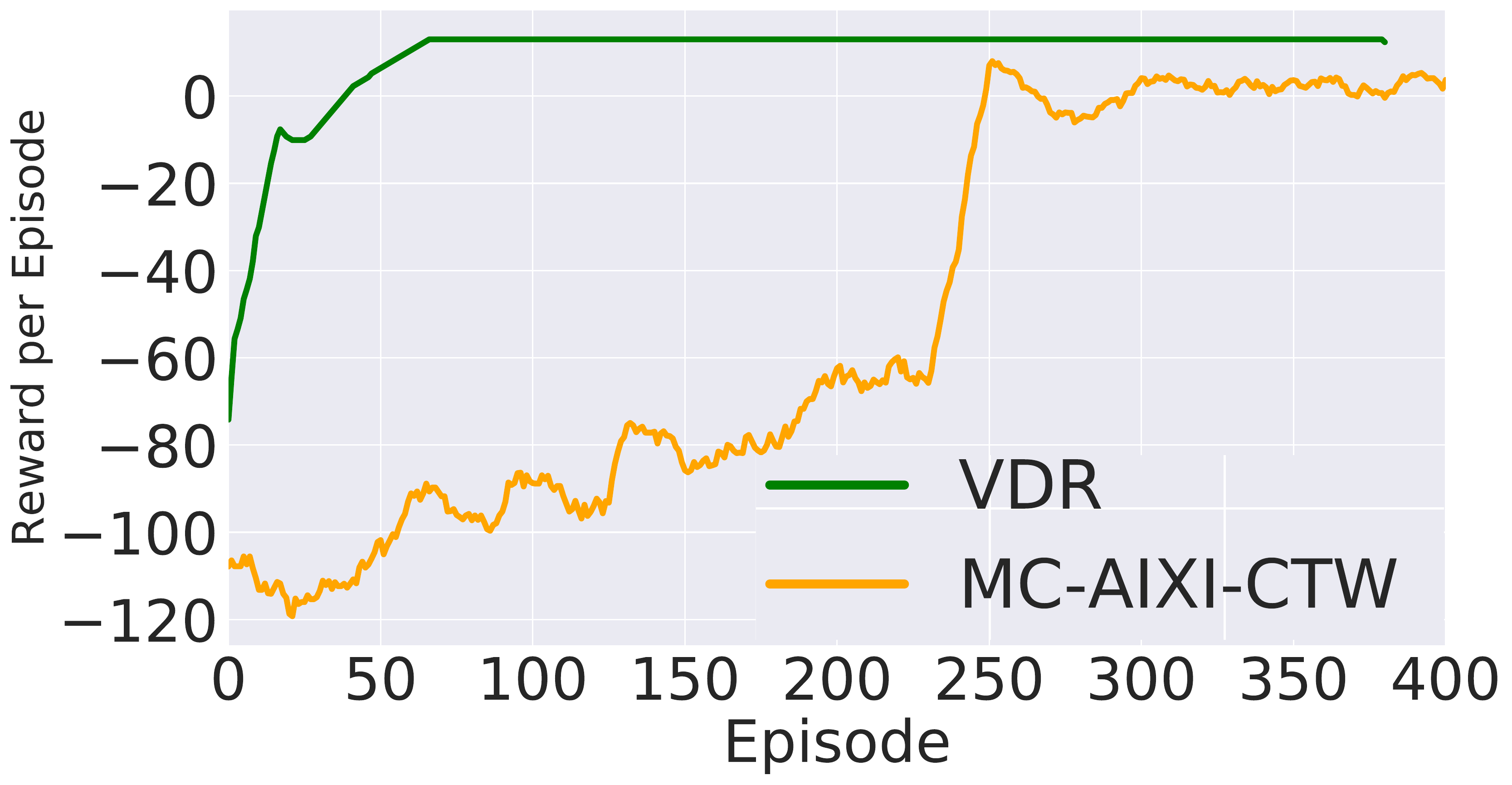}
  \caption{}
  \label{fig:CMComp}
\end{subfigure}
\caption{Cheese Maze, (a) Environment, agent has 4 actions: move up, down, left, right and receives a reward of -10 if it hits the wall, -1 for moving to an empty state and 10 for getting to the cheese state which also teleports the agent to the initial state (bottom left). Numbers in each cell indicate the initial observation space, and split of observation 1 into (1a, 1b). (b) Comparison of our algorithm to MC-AIXI-CTW}
\label{fig:CM}
\end{figure}

\section{SIMULATION RESULTS}\label{sec:res}
This human-in-the-loop RL system is designed ultimately to be used for helping 
humans and RL agents best work together to achieve a good representation that 
can be used to quickly identify good policies. However, as an initial proof 
of concept, we first conduct simulated experiments where a simulated 
human designer will agree a proposed observation refinement is beneficial 
if $c=0.25$ and where the revised associated observation is split as 
follows.  Assuming MDP states $s \in \mathcal{S}_o$ was clustered under observation 
$o$, we assign state $s$ to observation $o^1$ (or $o^2$) if more than 50 percent 
of the time $s$ was assigned to $o^1$ (or $o^2$) by the Viterbi algorithm. 

We simulate our VDR algorithm in two existing RL tasks. One is a navigation task CheeseMaze~\citep{veness2011monte} 
inspired by a robot needing to learn how to reach a destination: the robot requires 
particular features in order to be able to learn and represent the optimal policy. 
The second is mountain car~\citep{sutton1998reinforcement}, where a car on a hill 
must reach the goal position up the right side of the hill (figure \ref{fig:MClayoput}).

The goal is to see if our method will propose the necessary augmentation to the simulated human to learn the optimal policy. Additionally, we are interested to evaluate how fast our algorithm can learn starting from a coarse observation space, when representing the optimal policy does not require full Markovian state space.

\subsection{Cheese Maze}

Cheese Maze was used as a benchmark environment in \cite{veness2011monte}; for details refer to figure \ref{fig:CMlayout}. We set the maximum length of each episode to 20, and consider an augmentation every 5 episodes. We compare to MC-AIXI-CTW that outperformed other history based and feature RL method including UTree~\citep{mccallum1996learning, veness2011monte}.

As shown in Figure \ref{fig:CMComp}
VDR outperforms MC-AIXI-CTW and finds the optimal policy in fewer number of episodes. 
VDR finds the optimal policy by splitting observation 1 (see figure \ref{fig:CMlayout}) into two (1a, 1b) that requires different action for representing the optimal policy in center and left side of the maze.

\subsection{Mountain Car}

\begin{figure}[tb]
\centering
\begin{subfigure}{.45\linewidth}
  \centering
  \includegraphics[width=\linewidth]{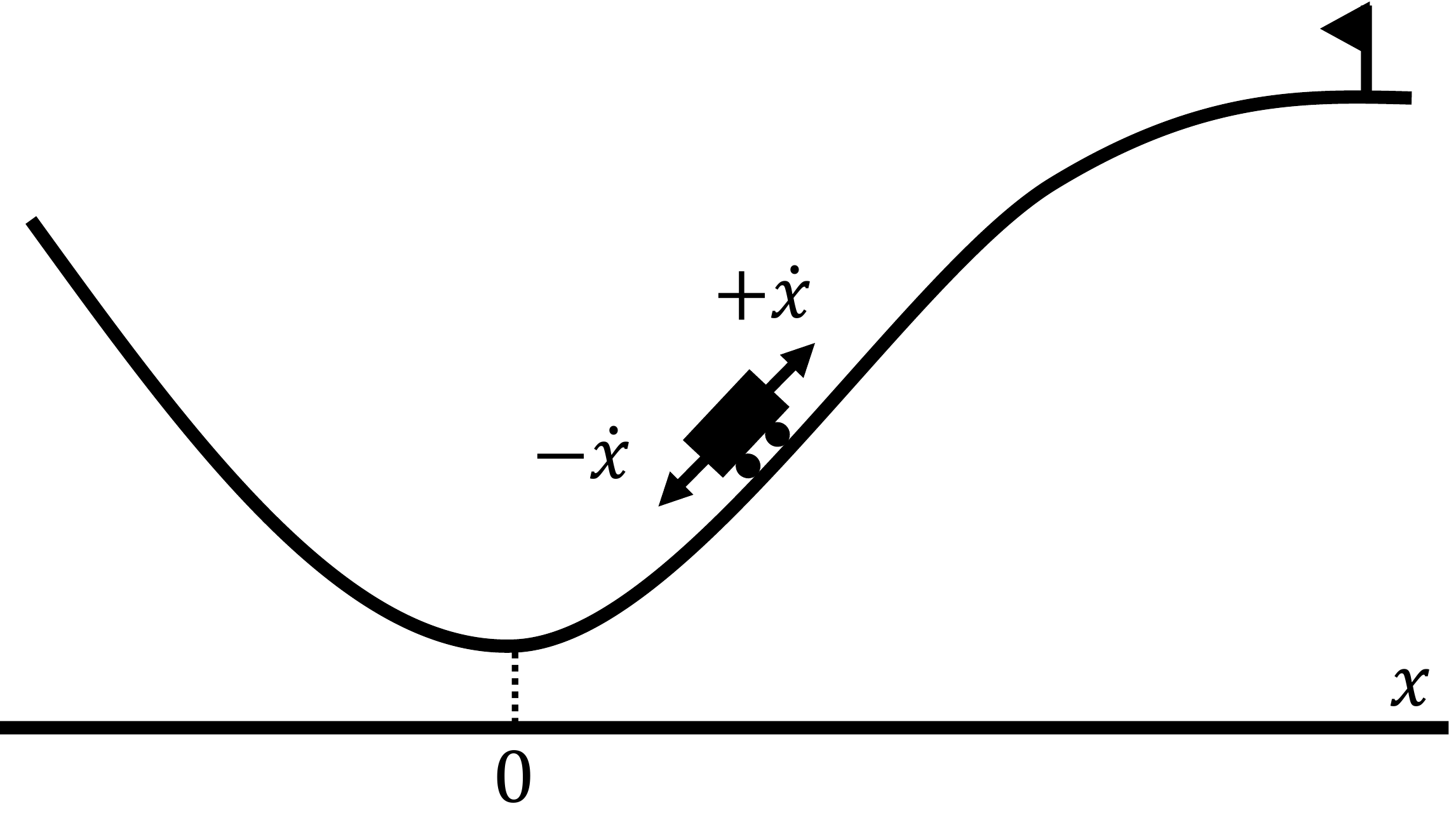}
  \caption{}
  \label{fig:MClayoput}
\end{subfigure}%
\quad
\begin{subfigure}{.45\linewidth}
  \centering
  \includegraphics[width=\linewidth]{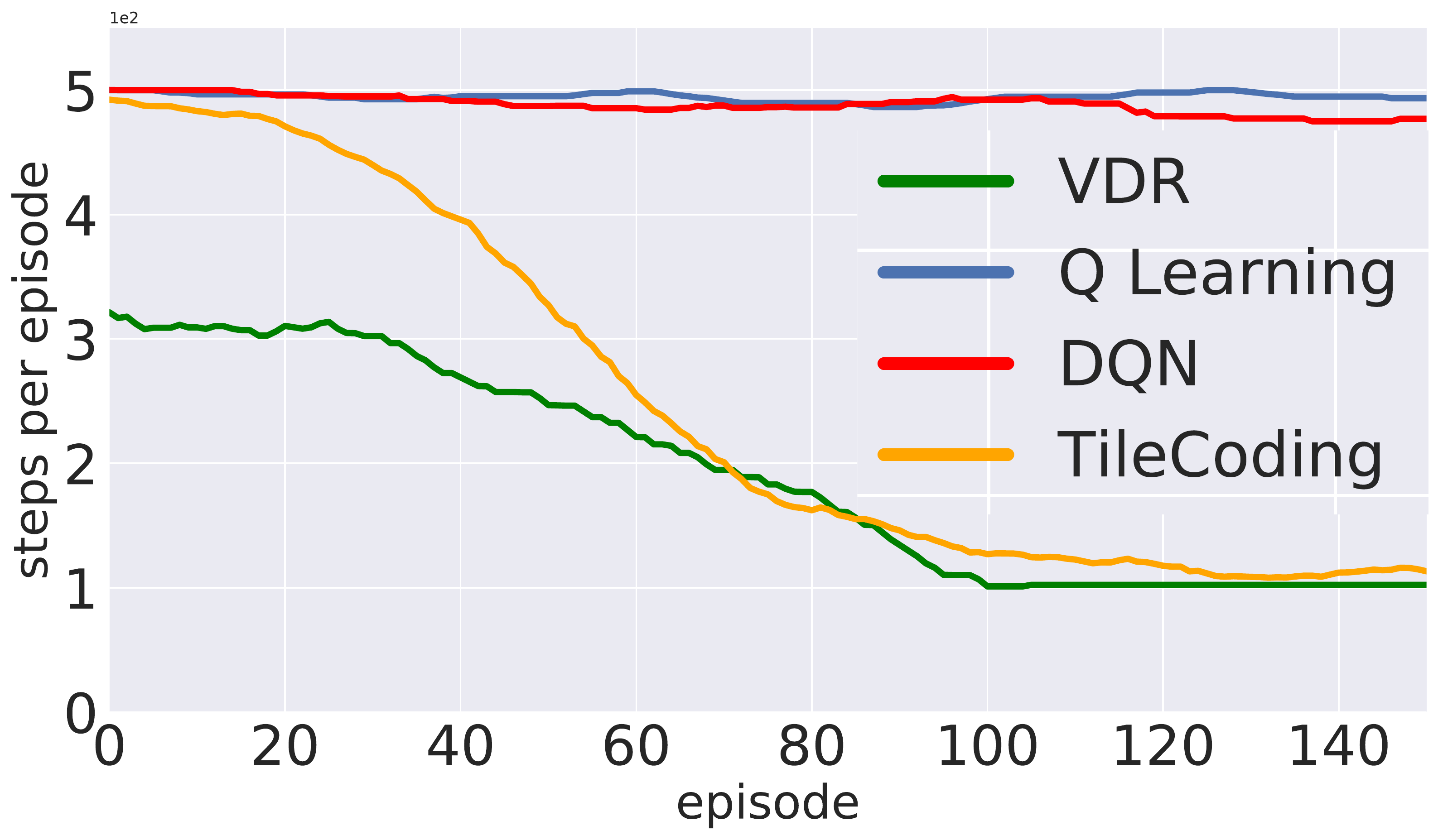}
  \caption{}
  \label{fig:MCComp}
\end{subfigure}

\begin{subfigure}{\linewidth}
  \centering
  \includegraphics[width=0.9\linewidth]{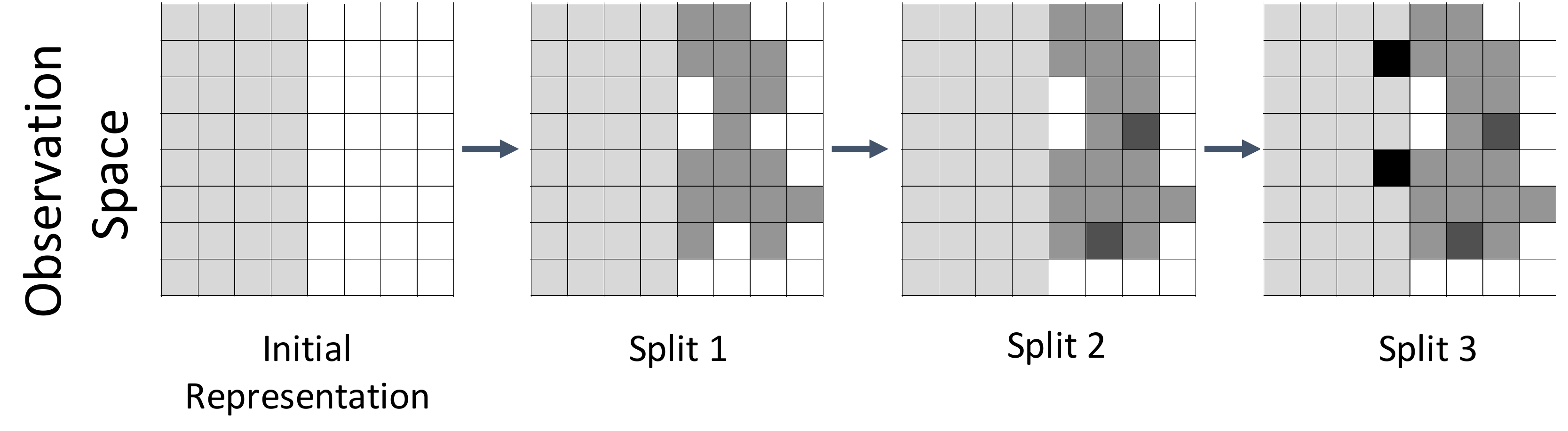}
  \caption{}
  \label{fig:MCObs}
\end{subfigure}

\caption{Mountain Car, (a) Environment. (b) Comparison of VDR with Q-learning, tile coding and DQN. (c) Observation space augmentation: each color represents distinct observation and the observation space is shown superimposed on the underlying 8x8 grid.}
\label{fig:CM}
\end{figure}


We considered mountain car where the agent always starts at the same location and velocity $x_0=-0.5, v_0=0.03$. In all simulations the initial starting position and velocity is fixed for that entire process. We set the maximum episode length to 500 and consider an observation augmentation every 20 episodes. We treat the underlying true state space as a discrete 8x8 grid, though the true space is best modeled continuously. We compared our algorithm with Q-learning with $\epsilon$-greedy exploration on 20x20 gird (location and velocity).
Additionally we compared to DQN~\cite{mnih2015human} (with two hidden layers of size 64) and tile coding with 2 tilings.

Figure~\ref{fig:MCComp} shows the result for $v_0 = 0.03$, where VDR can learn the optimal policy in ~100 episodes with only 3 augmentations: this is enough to represent the optimal policy. Our experiments with other initial velocity ($v_0=0.05$) showed the same results. Figure~\ref{fig:MCObs} shows the observation augmentation. Starting with only two observation shown in figure ~\ref{fig:MCObs} superimposed on the underlying 8x8 grid, VDR find the optimal policy after three splits.

\section{Future Work And Conclusion}
Adaptive interactive reinforcement learning systems often benefit from a 
human-in-the-loop system designer that can modify the state or action 
space of the RL system to improve performance. Given the potential set 
of modifications, guidance from the RL system to the human designer 
could be helpful. In this short paper we propose a way for a RL system 
to proactively propose augmentations to its observation representation that 
may enable improved performance. Our simulations suggest the potential 
benefit of this approach in small domains. Exploring the scalability of 
this approach and testing it with real humans in the loop 
are clear interesting next steps.

\section*{Acknowledgement}
The research reported here was supported by NSF BIGDATA award and ONR Young Investigator award.

\bibliographystyle{ACM-Reference-Format}
\bibliography{acmart}

\appendix
\section{Details of VDR Algroithm}
In this section we present some details of VDR algorithm (algorithm  1). Section \ref{app:opto} describes the details of off-policy policy optimization, section \ref{app:split} describes the details of augmenting the observation space and section \ref{app:traj} presents a running example of a trajectory tree.

\subsection{Off-Policy Policy Optimization}\label{app:opto}
A natural question is how can we update the policy parameters when using OPTO? By using the trajectory tree $\mathcal{T}$ as a simulator, any policy optimization method can be applied by performing roll-outs on the trajectory tree $\mathcal{T}$ to achieve a (near) optimal stochastic policy. For example, if the policy is parametrized by $\theta$, one can use
REINFORCE~\citep{williams1992simple} to update the parameters using $\theta_{t+1} = \theta_{t} + \alpha \nabla J(\theta)$, where $\nabla J(\theta)$ is
$$\nabla J(\theta) = \mathbb{E}_{\pi}[G_t \frac{\nabla_{\theta}\pi(a|s;\theta)}{\pi(a|s;\theta)}]$$

Where $G_t$ is the discounted return of an episode and the expectation can be calculated with a sample episode on the trajectory tree $\mathcal{T}$, as described in algorithm \ref{alg:reinforce}.

\begin{algorithm}
\caption{REINFORCE on $\mathcal{T}$}\label{alg:reinforce}
\begin{algorithmic}[1]
{\small
\REPEAT
\STATE Generate an episode with $\pi_{\theta}$ on $\mathcal{T}$
\FOR{$t=0,\dots,T-1$}
    \STATE $G_t \gets$ return from step $t$
    \STATE $\theta \gets \theta + \alpha \gamma^t G_t \nabla_{\theta} \log \pi(a_t|s_t;\theta)$
\ENDFOR
\UNTIL{timeout}
}
\end{algorithmic}
\end{algorithm}

\subsection{Splitting}\label{app:split}

In order to split (augment) an observation $o \in \mathcal{O}$, we notice that:

1) A predicted improvement in the value could arise due
to more accurate estimates of transition probabilities and rewards. Therefore a split
is only done if the best policy for the augmented observation
space is sufficiently different from the best policy before the split. In order to compare the two policies, we augment the policy before split $\pi_{old}$ to $\pi^{*}_{old}$ by setting $\forall a\in \mathcal{A}: \pi_{old}(o_i,a)=\pi_{old}^{*}(o_i^1,a)=\pi_{old}^{*}(o_i^2,a)$, and the same policy for all other observations. 

2) Observation augmentation will turn one observation into two,
and by definition indicates that at least one observation will
now have less counts in observed trajectories compared to
when that observation was not refined. This reduced data will
generally increase the variance of the computed estimated values
of the possible refined observation spaces. Therefore we use a
bootstrap procedure on the all old data to compute $B$
estimates of the refined observation values. In deciding whether
to split, we compare the potential benefit to the estimated
standard deviation across bootstrap estimates of the
value of the new proposed observation splits as
$$score(i) = D_{KL}(\pi^*_{old}||\pi_i)\frac{avg(V^{\pi_i}) - V^{\pi_{old}}}{std(V^{\pi_i})}$$
Where $D_{KL}$ is the KL-divergence. We split if $score$ exceeds an input threshold. This is a
heuristic estimate of a significance test (Z-score) 
for whether the algorithm is confident that the new split
representation will outperform the prior. Pseudo code for the complete splitting procedure is shown in Algorithm~\ref{alg:split}.

\begin{algorithm}[h!]
\caption{Split($\pi_{old}, graph, B=10, N=100$)}\label{euclid}
\label{alg:split}
\begin{algorithmic}[1]
\STATE $v_{old} \gets OPTE(\pi_{old}, graph,N,False)$
\STATE $\pi^*_{old} \gets$ augment the policy $\pi_{old}$
\FOR{$o_j \in \mathcal{O}$}
\STATE $o_j^1,o_j^2 \gets o_j$ // use EM to split observation
\STATE $\mathcal{O}_j \gets (\mathcal{O} \setminus o_j) \cup (o_j^1,o_j^2)$
\STATE Construct $graph_{j}$ for  $\mathcal{O}_j$
\STATE $\pi_j=\max_{\pi}$ OPTE($\pi,graph_j,N,False$) 
\FOR{$b=1:B$}
\STATE Construct $graph_j^b$ for bootstrap $b$, $\mathcal{O}_j$
\STATE $V^{\pi_j}(b)=$ OPTE($\pi_j,graph_j^b,N,False$) 
\ENDFOR
        \STATE $splitScore(j) = D_{KL}(\pi^{*}_{old}||\pi_j) \frac{avg(V^{\pi_j}) - v_{old}}{std(V^{\pi_j})}$
\ENDFOR
\IF{$\max(splitScore) > thresh$}
\STATE $J \gets argmax_j$ $splitScore$
\STATE $\mathcal{O} \gets
\mathcal{O}_{J}$
\ENDIF
\end{algorithmic}
\end{algorithm}

\subsection{Example of Trajectory Tree}\label{app:traj}
In order to make trajectory tree $\mathcal{T}$ clear, here is a concrete example of an environment and a trajectory associated with it. 

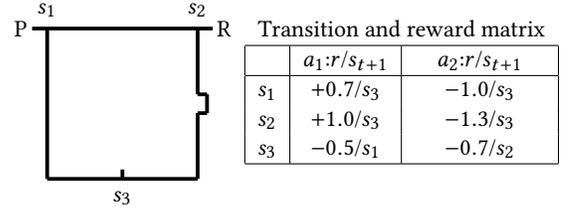
\begin{figure}[h]
\centering
  \begin{tikzpicture}[baseline=1.1cm]
  \draw [line width=0.5mm] (0,0) -- (2,0);
  \draw [line width=0.5mm] (1,0) -- (1,0.125);

  \draw [line width=0.5mm] (2,0) -- (2,1.75/2);
  \node[] at (2/2, -0.5/2) {$s_3$};

  \draw [line width=0.5mm] (4/2,1.75/2) -- (4.25/2,1.75/2);

  \draw [line width=0.5mm] (4.25/2,1.75/2) -- (4.25/2, 2.25/2);
  \draw [line width=0.5mm] (4.25/2,2.25/2) -- (4/2, 2.25/2);
  \draw [line width=0.5mm] (4/2, 2.25/2) -- (4/2,4/2);

  \draw [line width=0.5mm] (0,0) -- (0,4/2);
  \draw [line width=0.5mm] (-0.4/2,4/2) -- (4.4/2,4/2);
  \node[] at (0/2,4.5/2) {$s_1$};
  \node[] at (4/2,4.5/2) {$s_2$};
  \node[] at (-0.7/2,4/2) {P};
  \node[] at (4.7/2,4/2) {R};
  \end{tikzpicture}
  \begin{tabular}{|c|c|c|}
    \multicolumn{3}{c}{Transition and reward matrix} \\
  		\hline
          \  & $a_1$:$r$/$s_{t+1}$ & $a_2$:$r$/$s_{t+1}$ \\ 
          \hline
          $s_1$ & $+0.7$/$s_3$ & $-1.0$/$s_3$ \\ 
          $s_2$ & $+1.0$/$s_3$ & $-1.3$/$s_3$ \\
          $s_3$ & $-0.5$/$s_1$ & $-0.7$/$s_2$ \\
          \hline
     \end{tabular}
\caption{3-state MDP introduced by \cite{mccallumThesis}. Agent can take an action right ($a_1$) or left ($a_2$) from any T-intersection, agent is always facing the wall at any intersection and starts at $s_3$. Action from $s_1$ and $s_2$ immediately teleports to state $s_3$. Immediate reward is sum of two components: some negative reward for walking plus reward through reward (R) or punishment (P) gate.}
\label{fig:3state}   
\end{figure}

We use a 3-state deterministic Markov decision process introduced by \cite{mccallumThesis} (see Figure~\ref{fig:3state} for full details). Consider that at the start the initial observation space $\mathcal{O}_0$ aliases all 3 states into a single observation $o_1$. Also assume that each episode lasts for 3 time steps, and an agent has acted in this decision process for 2 episodes where it only has access to the observations space (and not the true states). Let this initial data be 
as defined in Table~\ref{tab:data} which displays both the true 
latent states ($\mathcal{S}$) and the observations ($\mathcal{O}_0$) available to the agent.

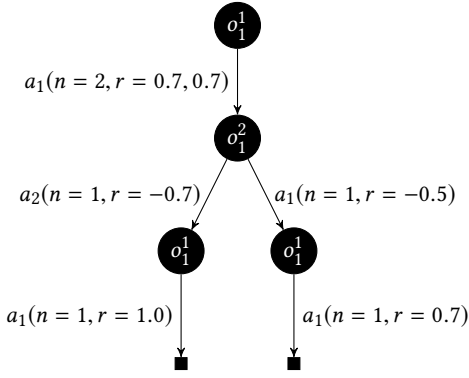
\begin{figure}
\centering
\begin{tikzpicture}[->,>=stealth',level/.style={sibling distance = 3cm/#1,
  level distance = 1.5cm}] 
\node [arn_n] {$o_{1}^{1}$}
    child{ node [arn_n] {$o_{1}^{2}$} 
         child{ node [arn_n] {$o_{1}^{1}$} 
            child{ node [arn_x] { } edge from parent node[left] {$a_1(n=1, r=1.0)$}
         }edge from parent node[left] {$a_2(n=1, r=-0.7)$}
    } 
    child{ node [arn_n] {$o_{1}^{1}$}
							child{ node [arn_x] { } edge from parent node[right]{$a_1(n=1,r=0.7)$} }edge from parent node[right]{$a_1(n=1,r=-0.5)$}
    } edge from parent node[left] {$a_1(n=2,r=0.7,0.7)$}                 
    }; 
\end{tikzpicture}
\caption{Trajectory Tree $\mathcal{T}$ of 2 episodes for 3-state MDP after 1 observation split, for the data shown in Table~\ref{tab:data}. Actions are represented by edges which include the count $n$ of the number of times the action has been taken given its ancestor nodes, and the rewards obtained during those experiences. Nodes are labeled with observations. Here as the dynamics are deterministic, each action goes to a single next observation, but more generally there can be observation branching for each action.}
\label{fig:grapho}
\end{figure}

Here there is only 1 observation and only 1 potential split. 
While EM is a procedure only guaranteed to yield a local optima, 
in this case one such optima would be that states 
$s_1$ and $s_2$ remain aliased to a single observation $o_{1}^{1}$ but 
state $s_3$ is distinguished and represented as observation $o_{1}^{2}$, 
as illustrated in Table~\ref{tab:data}. Given this split, Figure \ref{fig:grapho} shows the trajectory tree $\mathcal{T}$ obtained by data shown in table \ref{tab:data}.

\begin{table}[h]
    \centering
\begin{tabular}{|c|c|c|c|}
\multicolumn{4}{c}{Trajectory 1}\\
\hline
$\mathcal{S}$ & ($s_1$, $a_1$, 0.7) & ($s_3$, $a_2$, -0.7) & ($s_2$, $a_1$,+1.0) \\ \hline
$\mathcal{O}_0$  & ($o_{1}$, $a_1$, 0.7) & ($o_{1}$, $a_2$, -0.7) & ($o_{1}$, $a_1$,+1.0) \\ \hline
$\mathcal{O}_1$ & ($o_{1}^{1}$, $a_1$, 0.7) & ($o_{1}^{2}$, $a_2$, -0.7) & ($o_{1}^{1}$, $a_1$,+1.0) \\ \hline
\multicolumn{4}{c}{\ }\\
\multicolumn{4}{c}{Trajectory 2}\\
\hline
$\mathcal{S}$ & ($s_1$, $a_1$, 0.7) & ($s_3$, $a_1$, -0.5) & ($s_1$, $a_1$, +0.7) \\ \hline
$\mathcal{O}_0$ & ($o_{1}$, $a_1$, 0.7) & ($o_{1}$, $a_1$, -0.5) & ($o_{1}$, $a_1$, +0.7) \\ \hline
$\mathcal{O}_1$ & ($o_{1}^{1}$, $a_1$, 0.7) & ($o_{1}^{2}$, $a_1$, -0.5) & ($o_{1}^{1}$, $a_1$, +0.7) \\ \hline
\end{tabular}
    \caption{2 episodes of 3 time steps from 3-state MDP. $\mathcal{S}$ denotes the Markov  sequence of latent states, actions and rewards, $\mathcal{O}_0$ shows the sequence observed to the agent given an observation space that aliases all the states, $\mathcal{O}_1$ shows the sequence observed to the agent  after the split of $o_1$ into $o_{1}^{1}$ and $o_{1}^2$.}
    \label{tab:data}
\end{table}

\section{ASYMPTOTIC ANALYSIS}
In this section we analyze the asymptotic behaviour of our proposed algorithm, under the assumptions of infinite data. This assumption allows us to study the asymptotic behaviour, however as shown in experiments, our empirical evaluation shows the desired effect with limited data.  By this assumption we can have an accurate estimate of $V^{\pi}$ using OPTE by setting $C_1 = 0$ and $opt$ to $False$. 

\begin{lemma}\label{lemma:1}
    Let $M$ be a Markov decision process, and $\mathcal{T}$ be a trajectory tree generated by infinite data gathered using a policy $\pi\ s.t.\ \forall o,a: \pi(o,a) > 0$. Then $\forall \pi: \mathcal{O} \times \mathcal{A} \rightarrow [0,1]$, $V_{\mathcal{T}}^{\pi} \xrightarrow{i.p.} V_M^{\pi}$. Where $V_{\mathcal{T}}^{\pi}$ and $V_{M}^{\pi}$ are values of the policies evaluated in trajectory tree $\mathcal{T}$ and MDP $M$, respectively.
\end{lemma}
\begin{proof}
    This follows by the fact that with infinite data evaluating the policy using $\mathcal{T}$ doesn't require bootstrapping $Q$ values at the leaf, and all the history based transition probabilities and rewards converges in probability to their real values by the law of large numbers. As a result value of the policy, $\sum_{t \sim \pi} p(t) G(t)$ converges in probability to its value in MDP $M$, where $t \sim \pi$ are all the trajectories generated by policy $\pi$, $p(t)$ is the probability of trajectory under policy $\pi$, and $G(t)$ is the return of the trajectory.
\end{proof}

However, in the case of limited data the accuracy of $V_{\mathcal{T}}^{\pi}$ depends on the accuracy of the model free $Q(o,a)$ estimates.

\begin{theorem}\label{theo:markov}
Let $M$ and $\tilde{M}$ be Markov decision processes over the same action space $\mathcal{A}$ and state spaces $\mathcal{S}$, $\tilde{\mathcal{S}}$, respectively. Where $\tilde{\mathcal{S}} = \mathcal{S} \setminus s_0 \cup \{s^1_0,s^2_0\}$ such that $s_0^1, s_0^2$ are the split of state $s_0$. Let $\pi^* : \mathcal{S} \rightarrow \mathcal{A}$ be the optimal policy in $M$. Then, $\exists \tilde{\pi}: \tilde{\mathcal{S}} \rightarrow \mathcal{A}$ such that $V_{\tilde{M}}^{\tilde{\pi}} = V_{M}^{\pi}$ where $\tilde{\pi}(s_0^1) = \tilde{\pi}(s_0^2) = \pi(s_0)$, and $\forall \pi' \neq \tilde{\pi} : V_{\tilde{M}}^{\pi'} \leq V_{\tilde{M}}^{\tilde{\pi}}$.
\end{theorem}

\begin{proof}
Since both $M$ and $\tilde{M}$ are MDP, W.L.G we assume that all polices are deterministic. Setting $\tilde{\pi}(s_0^1) = \tilde{\pi}(s_0^2) = \pi^*(s_0)$ will trivially retrieve the optimal policy of $M$ with the same value $V_{\tilde{M}}^{\tilde{\pi}} = V_{M}^{\pi}$. Now consider two cases when :
\begin{enumerate}
    \item $\pi'(s_0^1) = \pi'(s_0^2)$: then $\pi'$ is also a policy in $M$ with $\pi'(s_0^1) = \pi'(s_0^2) = \pi'(s_0)$, where $V_M^{\pi'} \leq V_M^{\pi^*} = V_{\tilde{M}}^{\tilde{\pi}}$, since $\pi^*$ is the optimal policy in $M$. 
    \item $\pi'(s_0^1) \neq \pi'(s_0^2)$: W.L.G assume $V^{\pi'}(s_0^2) \geq V^{\pi'}(s_0^1)$, we show that the policy can be improved by setting $\pi'(s_0^1)$ equal to $\pi'(s_0^2)$. Then followed by case 1, $V_M^{\pi'} \leq V_M^{\pi^*} = V_{\tilde{M}}^{\tilde{\pi}}$.
\end{enumerate}
\begin{align*}
    &Q^{\pi'}(s_0^1, \pi'(s_0^2)) \\&= R(s_0^1, \pi'(s_0^2)) + \gamma \sum_{s'}p(s'|s_0^1, \pi'(s_0^2)) V^{\pi'}(s')\\
    &= R(s_0^2, \pi'(s_0^2)) + \gamma \sum_{s'}p(s'|s_0^2, \pi'(s_0^2)) V^{\pi'}(s')\\
    &= Q^{\pi'}(s_0^2, \pi'(s_0^2))
    = V^{\pi'}(s_0^2) \geq V^{\pi'}(s_0^1) = Q^{\pi'}(s_0^1, \pi'(s_0^1))
\end{align*}

The second equality comes from the fact that in Markovian representation, state is sufficient to determine the transition probabilities and rewards so,
\begin{align*}
    \forall s,a: P(s|s_0^1,a) &= P(s|s_0^2,a)\\
    \forall a: R(s_0^1, a) &= R(s_0^2,a)
\end{align*}
\end{proof}

Theorem \ref{theo:markov} states that, by Lemma \ref{lemma:1}, our algorithm will not split a Markov representation further. However, in the case of limited data and using policy optimization to find the optimal policy, we might not get an accurate estimate of the value, or find the deterministic policy that is optimal in MDP. 

The $\pi^*$-irrelevance abstraction~\citep{li2006towards}  is a state abstraction of an MDP that, in every cluster, the optimal action is the same. The following lemma states that our algorithm will not split a observation space that is a $\pi^*$-irrelevance abstraction.

\begin{lemma}
If the observation space $\mathcal{O}$ is $\pi^*-irrelevance$ abstraction of an MDP $M$, our algorithm will not split further.
\end{lemma}
\begin{proof}
    (sketch) Based on theorem \ref{theo:markov}, the optimal policy is the optimal policy in Markovian state representation, and the optimal policy in $\pi^*-irrelevance$ abstraction has the same value as the optimal policy for $M$. Thus, there does not exist a split that yields higher return and our algorithm will not split further. 
\end{proof}

%
%

\end{document}